\documentclass[conference]{IEEEtran}
\IEEEoverridecommandlockouts
\usepackage{siunitx}

\ExplSyntaxOn
\msg_redirect_name:nnn
  { siunitx }
  { option-preamble-only }
  { none }
\ExplSyntaxOff
\usepackage{cite}
\usepackage{amsmath,amssymb,amsfonts}
\usepackage{algorithmic}
\usepackage{graphicx}
\usepackage{textcomp}
\usepackage{xcolor}
\usepackage[subpreambles=true]{standalone}
\usepackage{graphicx}
\usepackage{amsmath}
\usepackage{amssymb}
\usepackage{booktabs}
\usepackage{import}
\usepackage{times}
\usepackage{epsfig}
\usepackage{graphicx}
\usepackage{amsmath}
\usepackage{amssymb}
\usepackage{booktabs}
\usepackage{float}
\usepackage{xr}
\usepackage[pagebackref,breaklinks,colorlinks]{hyperref}

\newcommand{\etal}{et~al.\ }

\usepackage[capitalize]{cleveref}
\def\BibTeX{{\rm B\kern-.05em{\sc i\kern-.025em b}\kern-.08em
    T\kern-.1667em\lower.7ex\hbox{E}\kern-.125emX}}

\begin{document}

\title{StaticSegFormer: An Efficient\\ High-Performance Semantic Segmentation\\ Based on Static Structured Pruning}

\author{
\IEEEauthorblockN{
Timo Bartels$^{1}$,
Danish Nazir$^{1,2}$,
Jan Piewek$^{2}$,
Thorsten Bagdonat$^{2}$,
Tim Fingscheidt$^{1}$
}
\IEEEauthorblockA{
$^{1}$ Technische Universität Braunschweig, Germany\\
{\tt\small \{timo.bartels,danish.nazir,t.fingscheidt\}@tu-bs.de}\\
$^{2}$ Group Innovation, Volkswagen AG, Wolfsburg, Germany\\
{\tt\small \{danish.nazir,jan.piewek,thorsten.bagdonat\}@volkswagen.de}
}
}

\maketitle

\begin{center}
\small
This work has been submitted to the IEEE for possible publication.
Copyright may be transferred without notice, after which this version
may no longer be accessible.
\end{center}

\begin{abstract}
Structured pruning enhances the efficiency of deep neural networks (DNNs) by eliminating groups of parameters during inference. Previous methods mostly reduce computational complexity (FLOPs), while semantic segmentation performance (mIoU) slightly drops. Accordingly, recent dynamic structured pruning methods aim at reducing the performance drop, while lowering the FLOPs even more. However, on the ADE20K and Cityscapes benchmarks, our study reveals that on a GPU platform such dynamic methods exhibit a surprisingly low frame rate far below a simple static approach, while having comparable results in mIoU and FLOPs. To address this issue, we propose a static structured pruning method for attention layers, that achieves both, a lower FLOPs and a high frame rate [fps] of the \texttt{SegFormer} network, the latter increased by up to $34\%$ relative on the Cityscapes dataset, while having no mIoU performance drop at all. Our so-called \texttt{StaticSegFormer} method is strongest for small encoders and large images.
\end{abstract}


\section{Introduction}
Since the advent of deep convolutional neural networks (CNNs),
computer vision has advanced significantly.
In semantic segmentation, which assigns class labels to individual pixels,
CNNs have outperformed other methods on numerous benchmarks.
However, since the introduction of vision transformers (ViTs)
by Dosovitskiy \etal~\cite{Dosovitskiy2021},
ViTs have come to dominate most semantic segmentation benchmarks. In our work, due to its high performance and efficiency, we build upon the ViT \texttt{SegFormer} \cite{Xie2021}. Xie \etal \cite{Xie2021} improved upon the original ViT \cite{Dosovitskiy2021} by proposing a more efficient attention module, a lightweight decoder, and a hierarchical encoder. Nevertheless, despite these enhancements, their proposed attention layer still imposes a high computational demand, which scales quadratically with the number of input pixels.

Floating-point operations (FLOPs) are the most widely used metric for reporting the computational efficiency of deep neural networks (DNNs), enabling standardized, hardware-independent comparisons across architectures. However, FLOPs often correlate poorly with actual speed or energy consumption on the target hardware, limiting their practical relevance. We therefore also report frames per second (fps) on an \texttt{NVIDIA A100} and a \texttt{GTX 1080 Ti}.

To increase the efficiency of DNN structured pruning, a special form of pruning, where entire model structures, such as kernels, neurons, or layers are removed, can be applied to reduce FLOPs and increase fps substantially \cite{Bai, CAP, slimming, ACOSP}. However, most structured pruning methods trade efficiency (fps, FLOPs) with a certain loss in mIoU performance. Recent structured pruning approaches proposed to prune model structures dynamically based on the layer input, therefore keeping the original model capacity, while reducing the mIoU performance drop \cite{Bai,gao2018dynamic}. Despite these efforts, dynamic structured pruning still incurs
mIoU losses. Moreover, we find that Bai \etal~\cite{Bai}'s
approach is slower than its unpruned counterpart, as dynamically
assembling matrices in linear layers introduces computational
overhead exceeding the savings from reduced dimensions.

This is why in this work, we aim at finding a \textit{static} structured pruning approach, that achieves on the one hand a \textit{higher efficiency} in terms of \textit{both} FLOPs and fps, while at the same time obtaining a \textit{higher mIoU}. To achieve this, we build upon \texttt{DynaSegFormer} \cite{Bai}, but apply structured pruning solely to the attention layers. We demonstrate that such simple yet effective static structured pruning of the heads in the multi-head attention layers yields comparable results to the dynamic gating module and pruning ratio annealing as proposed by Bai \etal \cite{Bai}, both in terms of FLOPs and mIoU. Our method does not suffer from the drawbacks regarding inference time, instead, it significantly increases the fps without the need of additional techniques, such as knowledge distillation utilized by Bai \etal \cite{Bai}. Further, we extensively study pruning ratios across different
\texttt{SegFormer} encoder sizes to determine whether a heavily
pruned large encoder can outperform a less pruned smaller one.
Finally, we evaluate the best-performing networks on
Cityscapes~\cite{Cordts2016} and ADE20K~\cite{Zhou2017},
demonstrating our method's superiority over both
\texttt{SegFormer}~\cite{Xie2021} and
\texttt{DynaSegFormer}~\cite{Bai}.

\section{Related Work}
\subsection{Semantic Segmentation}
Semantic segmentation is a pixel-wise classification task, where each pixel of the input image is classified into predefined classes. With the rise of fully convolutional deep neural networks \cite{Long2015, Noh2015}, dense predictions became feasible with reasonable computational demands, achieving state-of-the-art performance on various benchmarks \cite{Cordts2016, Lin2014microsoft,Zhou2017}. Afterwards, Ronneberger \etal \cite{Ronneberger2015}
introduced \texttt{UNet}, combining deep and shallow features
to integrate global semantic context with local object boundary
information, an approach underlying most modern architectures.
Vision transformers, introduced by Dosovitskiy \etal \cite{Dosovitskiy2021},
now dominate state-of-the-art semantic segmentation
\cite{Cheng_2022_CVPR,10205465,Fang_2023_CVPR}.
Among various enhancements is the efficient self-attention
mechanism of \texttt{SegFormer} \cite{Xie2021},
on which we build in this work.

\subsection{Efficient Semantic Segmentation}
Most works focusing on enhancing network architecture to improve efficiency, as it has the most significant impact on model performance. Howard \etal \cite{Howard2017} introduced depth-wise separable convolution to reduce computational demand. Tan \etal \cite{tan2019efficientnet} proposed an efficient model scaling method for convolutional neural networks. With the emergence of vision transformers, various works have proposed more efficient attention mechanisms \cite{Xie2021, Cheng_2022_CVPR,Xia_2022_CVPR}. However, besides altering network structure, several other approaches exist, such as knowledge distillation \cite{10204863, Hinton2014, 9578915, Romero2015}, quantization \cite{8953869, Nagel_2019_ICCV, Fang2020Post}, and model pruning \cite{CAP, ACOSP, 10378640, Bejnordi2020Batch-shaping, Bai}, which can be additionally applied to improve efficiency.


\textbf{Model Pruning}: One approach to reducing the number of parameters in a model while maintaining mIoU performance at high level is through model pruning \cite{han2015learning, slimming, varghese2022joint,frankle2018the, fingscheidt_dnndataautomateddriving}. Pruning typically involves multiple stages, including pretraining, pruning, and final fine-tuning \cite{slimming}. Often, an L1 or L2 regularization term is employed to encourage network weights to converge towards zero, enabling their pruning without significant impact on mIoU performance \cite{slimming}. While unstructured pruning \cite{han2015learning, frankle2018the} results in sparse matrices and kernels, it fails to decrease model complexity during inference without specialized hardware capable of sparse matrix multiplication.

\textbf{Structured Model Pruning}: Structured pruning can be utilized to learn more efficient models in terms of FLOPs and fps \cite{slimming, CAP, ACOSP} as entire network structures converge to have minimal impact and can subsequently be pruned away.
Previous works have primarily focused on image classification.
Lebedev \etal \cite{Lebedev16} proposed to prune the kernel size of individual filters in a single convolutional layer. He \etal \cite{he2019filter} identified redundant filters in convolutional layers and pruned entire output channels. Similarly, Liu \etal \cite{slimming} applied L1 regularization to the batch normalization scaling parameter to prune entire convolutional output channels.
Michel \etal \cite{NEURIPS2019_2c601ad9} pruned entire attention heads in a vision transformer, showing minimal performance drop. Tang \etal \cite{Tang_2022_CVPR} masked out patch embeddings in vision transformers, thereby reducing redundant patches in later layers.
Semantic segmentation poses a challenge for structured pruning due to dense predictions. Nonetheless, some works exist. He \etal \cite{CAP} followed a similar approach to Liu \etal \cite{slimming}, utilizing batch normalization layers to prune convolutional channels, but additionally use their similarity as guidance. Ditschuneit \etal \cite{ACOSP} employed a temperature annealing schedule to prune a predefined set of convolutional channels.

\textbf{Dynamic Model Pruning}:
Classical structured pruning identifies and encourages the emergence of irrelevant model structures through an importance metric and regularization term with the goal of removing them entirely for faster inference. However, this often leads to a reduction in model capacity and a subsequent performance drop. Therefore, a new type of structured pruning has emerged, involving the dynamic reduction of the computational demand of a network based on the current input by switching on and off model structures.
Rao \etal \cite{DynamicVIT} dynamically prune vision transformer
patches for classification using an MLP on each patch embedding.
Gao \etal \cite{gao2018dynamic} dynamically prune convolutional
channels based on relevance predicted from preceding layer outputs.
For semantic segmentation, Tang \etal \cite{10378640} proposed dropping patch embeddings in vision transformers that already have high confidence, with auxiliary heads connected to earlier layers.
Bejnordi \etal \cite{Bejnordi2020Batch-shaping} dynamically switch off convolutional channels via a gating module and enforce the gate to be highly conditional on the input data through a method called batch-shaping. Bai \etal \cite{Bai} uses a dynamic multi-layer perceptron for self-attention to reduce the computational demand based on the layer input.

We build upon the latter dynamic approach from Bai \etal \cite{Bai}, as it is very effective in reducing computational complexity in terms of FLOPS. However, we show that using dynamic self-attention is not necessary, accordingly, we rely on static structured pruning. Our method reduces FLOPS even more than the \texttt{DynaSegFormer} from Bai \etal \cite{Bai} and is able to increase the frames per second (fps) compared to the \texttt{SegFormer} \cite{Xie2021}. We also build upon the structured pruning method from Michel \etal \cite{NEURIPS2019_2c601ad9} and prune attention heads, as it is highly successful for image classification. We show that such simple yet very effective static structured pruning method can be applied to a dense prediction task such as semantic segmentation. Different to previous works, we prune the attention heads directly after pretraining on ImageNet \cite{Deng2009}, thereby reducing training memory consumption and training time of the fine-tuning step. We demonstrate that \textit{our method enables the creation of highly efficient vision transformers that can be customized in size}, making them interesting for deployment on edge devices, as this optimizes the utilization of limited computing capacity, all without the need for repeating the expensive pre-training.

\section{Method}
\begin{figure}[t]
\begin{center}
\resizebox{0.75\columnwidth}{!}{
\begin{tikzpicture}
\coordinate  (start) at (0,0.3) {};
\node [fill=tu4, shape=rectangle, minimum width=9.0cm, minimum height=13.75cm, align=center] (BACK) at (-0.1,-6.9) {};
\node [inner sep=0,draw=black, thick, fill=white, shape=rectangle, minimum width=4.5cm, minimum height=0.58cm, align=center] (OPE1) at (0,-0.6) {Conv(7$\times$7, $F_1$)$_{/4}$};
\node [draw=black, thick, fill=white, shape=rectangle, minimum width=4.5cm, minimum height=0.58cm, align=center] (LN1) at (0,-1.25) {Layer Normalization};
\node [draw=black, thick, fill=tu63, shape=rectangle, minimum width=4.5cm, minimum height=0.58cm, align=center] (TB01) at (0,-1.9) {Transformer Block};
\node [draw=black, thick, fill=tu63, shape=rectangle, minimum width=4.5cm, minimum height=0.58cm, align=center] (TB1) at (0,-2.9) {Transformer Block};
\node [inner sep=0,draw=black, thick, fill=white, shape=rectangle, minimum width=4.5cm, minimum height=0.58cm, align=center] (OPE2) at (0,-4.1) {Conv(3$\times$3, $F_2$)$_{/2}$};
\node [draw=black, thick, fill=white, shape=rectangle, minimum width=4.5cm, minimum height=0.58cm, align=center] (LN2) at (0,-4.75) {Layer Normalization};
\node [draw=black, thick, fill=tu63, shape=rectangle, minimum width=4.5cm, minimum height=0.58cm, align=center] (TB2) at (0,-5.4) {Transformer Block};
\node [draw=black, thick, fill=tu63, shape=rectangle, minimum width=4.5cm, minimum height=0.58cm, align=center] (TB2) at (0,-6.4) {Transformer Block};
\node [inner sep=0,draw=black, thick, fill=white, shape=rectangle, minimum width=4.5cm, minimum height=0.58cm, align=center] (OPE3) at (0,-7.6) {Conv(3$\times$3, $F_3$)$_{/2}$};
\node [draw=black, thick, fill=white, shape=rectangle, minimum width=4.5cm, minimum height=0.58cm, align=center] (LN2) at (0,-8.25) {Layer Normalization};
\node [draw=black, thick, fill=tu63, shape=rectangle, minimum width=4.5cm, minimum height=0.58cm, align=center] (TB03) at (0,-8.9) {Transformer Block};
\node [draw=black, thick, fill=tu63, shape=rectangle, minimum width=4.5cm, minimum height=0.58cm, align=center] (TB3) at (0,-9.9) {Transformer Block};
\node [inner sep=0,draw=black, thick, fill=white, shape=rectangle, minimum width=4.5cm, minimum height=0.58cm, align=center] (OPE4) at (0,-11.1) {Conv(3$\times$3, $F_4$)$_{/2}$};
\node [draw=black, thick, fill=white, shape=rectangle, minimum width=4.5cm, minimum height=0.58cm, align=center] (LN2) at (0,-11.75) {Layer Normalization};
\node [draw=black, thick, fill=tu63, shape=rectangle, minimum width=4.5cm, minimum height=0.58cm, align=center] (TB04) at (0,-12.4) {Transformer Block};
\node [draw=black, thick, fill=tu63, shape=rectangle, minimum width=4.5cm, minimum height=0.58cm, align=center] (TB4) at (0,-13.4) {Transformer Block};
\coordinate (end) at (0,-14.4) {} {} {} {} {} {} {} {} {} {} {} {} {} {} {} {} {} {} {};

\draw [-{Triangle[length=1.75mm,width=1.75mm]}] (start) -- (OPE1);

\draw [-{Triangle[length=1.75mm,width=1.75mm]}] (TB1) -- (OPE2);
\draw [-{Triangle[length=1.75mm,width=1.75mm]}] (TB2) -- (OPE3);
\draw [-{Triangle[length=1.75mm,width=1.75mm]}] (TB3) -- (OPE4);
\draw [-{Triangle[length=1.75mm,width=1.75mm]}] (TB4) -- (end);

\fill (0,-3.4) circle (2pt);
\draw [] (-3.4,-3.4) node (v1) {} -- (0,-3.4);
\draw [-{Triangle[length=1.75mm,width=1.75mm]}] (-3.4,-3.4) -- (-3.4,-14.4);

\fill (0,-6.9) circle (2pt);
\draw [] (-2.9,-6.9) node (v1) {} -- (0,-6.9);
\draw [-{Triangle[length=1.75mm,width=1.75mm]}] (-2.9,-6.9) -- (-2.9,-14.4);

\draw [] (-2.38,-10.4) node (v1) {} -- (0,-10.4);
\fill (0,-10.4) circle (2pt);
\draw [-{Triangle[length=1.75mm,width=1.75mm]}] (-2.38,-10.4) -- (-2.4,-14.4);

\node [rotate=0] (f1dim) at (0.8,0.15) {$H$$\times$$W$$\times$$3$};
\node [rotate=0] (f1dim) at (0.9,-3.5) {$\frac{H}{4}$$\times$$\frac{W}{4}$$\times$$F_1$};
\node [rotate=0] (f1dim) at (0.9,-7) {$\frac{H}{8}$$\times$$\frac{W}{8}$$\times$$F_2$};
\node [rotate=0] (f1dim) at (0.9,-10.5) {$\frac{H}{16}$$\times$$\frac{W}{16}$$\times$$F_3$};
\node [rotate=0] (f1dim) at (0.99,-14.1) {$\frac{H}{32}$$\times$$\frac{W}{32}$$\times$$F_4$};

\node [rotate=0] (f1dim) at (-0.2,0.2) {$\mathbf{x}$};

\node [rotate=0] (f1dim) at (-3.6,-14.0) {$\mathbf{f}_1$};
\node [rotate=0] (f1dim) at (-3.1,-14.0) {$\mathbf{f}_2$};
\node [rotate=0] (f1dim) at (-2.6,-14.0) {$\mathbf{f}_3$};
\node [rotate=0] (f1dim) at (-0.2,-14.0) {$\mathbf{f}_4$};

\node [rotate=0] (f1dim) at (3,-2.4) {$\times$$B_1$};
\node [rotate=0] (f1dim) at (3,-5.9) {$\times$$B_2$};
\node [rotate=0] (f1dim) at (3,-9.4) {$\times$$B_3$};
\node [rotate=0] (f1dim) at (3,-12.9) {$\times$$B_4$};

\node [rotate=0, align=left] (f1dim) at (-3.4,-0.7) {\texttt{SegFormer}\\\textbf{Encoder}};

\node [rotate=0] (f1dim) at (0,-2.3) {$\vdots$};
\node [rotate=0] (f1dim) at (0,-5.8) {$\vdots$};
\node [rotate=0] (f1dim) at (0,-9.3) {$\vdots$};
\node [rotate=0] (f1dim) at (0,-12.8) {$\vdots$};

\draw[thick,black,decorate,decoration={brace,amplitude=5pt}, transform canvas={yshift=-12pt}] (2.4,-1.2) -- node [below=12pt] {} (2.4,-2.8);
\draw[thick,black,decorate,decoration={brace,amplitude=5pt}, transform canvas={yshift=-12pt}] (2.4,-4.7) -- node [below=12pt] {} (2.4,-6.3);
\draw[thick,black,decorate,decoration={brace,amplitude=5pt}, transform canvas={yshift=-12pt}] (2.4,-8.2) -- node [below=12pt] {} (2.4,-9.8);
\draw[thick,black,decorate,decoration={brace,amplitude=5pt}, transform canvas={yshift=-12pt}] (2.4,-11.7) -- node [below=12pt] {} (2.4,-13.3);

\end{tikzpicture}}
\end{center}
   \caption{\textbf{Standard \texttt{SegFormer MiT} encoder}.
Varying transformer block counts $B_1,\dots,B_4$ and feature
dimensions $F_1,\dots,F_4$ yields the
\texttt{MiT-B0} to \texttt{MiT-B5} variants \cite{Xie2021}.
Our \texttt{StaticSegFormer} builds on this topology.}
\label{fig:segenc}
\end{figure}
\begin{figure}[t]
\begin{center}
\resizebox{5.0cm}{!}{
\begin{tikzpicture}
\coordinate  (start) at (0,2) {} {} {} {} {} {} {} {} {} {} {};
\node [fill=tu63, shape=rectangle, minimum width=7.0cm, minimum height=7.2cm, align=center] (BACK) at (-0.1,-1.7) {};
\node [draw=black, thick, fill=white, shape=rectangle, minimum width=4.5cm, minimum height=0.55cm, align=center] (LN1) at (0,0.5) {Layer Normalization};
\node [draw=black, thick, fill=white, shape=rectangle, minimum width=4.5cm, minimum height=0.55cm, align=center] (R1) at (0,-0.1) {Reshape};
\node [draw=black, thick, fill=tu113, shape=rectangle, minimum width=4.5cm, minimum height=0.55cm, align=center] (ESA) at (0,-1.2) {MHSA};
\node [draw=black, thick, fill=white, shape=rectangle, minimum width=4.5cm, minimum height=0.55cm, align=center] (R2) at (0,-1.8) {Reshape};
\node [draw=black, thick, fill=white, shape=rectangle, minimum width=4.5cm, minimum height=0.55cm, align=center] (LN2) at (0,-3.6) {Layer Normalization};
\node [draw=black, thick, fill=tu72, shape=rectangle, minimum width=4.5cm, minimum height=0.55cm, align=center] (MFFN) at (0,-4.2) {Mix-FFN};

\coordinate (end) at (0,-5.5) {} {} {} {} {} {} {} {};

\draw [-{Triangle[length=1.75mm,width=1.75mm]}] (start) -- (LN1);

\draw [-{Triangle[length=1.75mm,width=1.75mm]}] (R1) -- (ESA);

\fill (0,1.1) circle (2pt);
\draw [] (-2.5,1.1) node (v1) {} -- (0,1.1);
\draw [] (-2.5,1.1) node (v1) {} -- (-2.5,-2.6);
\draw [-{Triangle[length=1.75mm,width=1.75mm]}] (-2.5,-2.6) -- (-0.2,-2.6);
\draw [-{Triangle[length=1.75mm,width=1.75mm]}] (0,-2.1) -- (0,-2.4);
\draw [-{Triangle[length=1.75mm,width=1.75mm]}] (0,-2.8) -- (LN2);

\fill (0,-3) circle (2pt);
\draw [] (-2.5,-3) node (v1) {} -- (0,-3);
\draw [] (-2.5,-3) node (v1) {} -- (-2.5,-5);
\draw [-{Triangle[length=1.75mm,width=1.75mm]}] (0,-4.5) -- (0,-4.8);
\draw [-{Triangle[length=1.75mm,width=1.75mm]}] (-2.5,-5) -- (-0.2,-5);
\draw [-{Triangle[length=1.75mm,width=1.75mm]}] (0,-5.2) -- (end);

\node [rotate=0, align=left] (f1dim) at (-1.8,1.6) {\textbf{Transformer} \textbf{Block}};

\node[draw=black, thick, circle,fill=white,  minimum width=.4cm][] (adderBG2) at (0,-2.6) {};
\draw [] (0,-2.5) node (v1) {} -- (0,-2.7);
\draw [] (-0.1,-2.6) node (v1) {} -- (0.1,-2.6);

\node[draw=black, thick, circle,fill=white,  minimum width=.4cm][] (adderBG2) at (0,-5) {};
\draw [] (0,-4.9) node (v1) {} -- (0,-5.1);
\draw [] (-0.1,-5) node (v1) {} -- (0.1,-5);

\node [rotate=0] (f1dim) at (1.5,1.1) {$\frac{H}{4\cdot2^{j-1}}$$\times$$\frac{W}{4\cdot2^{j-1}}$$\times$$F_j$};
\node [rotate=0] (f1dim) at (1.4,-0.65) {$\frac{H}{4\cdot2^{j-1}}$$\cdot$$\frac{W}{4\cdot2^{j-1}}$$\times$$F_j$};
\node [rotate=0] (f1dim) at (1.6,-2.6) {$\frac{H}{4\cdot2^{j-1}}$$\times$$\frac{W}{4\cdot2^{j-1}}$$\times$$F_j$};
\node [rotate=0] (f1dim) at (1.6,-5) {$\frac{H}{4\cdot2^{j-1}}$$\times$$\frac{W}{4\cdot2^{j-1}}$$\times$$F_j$};

\end{tikzpicture}}
\end{center}

   \caption{\textbf{Transformer Block} that is employed in the standard \texttt{SegFormer} encoder and in our \texttt{StaticSegFormer} encoder as shown in \autoref{fig:segenc}.
   }
\label{fig:transformerblock}
\end{figure}
\subsection{\textbf{\texttt{(Dyna)SegFormer}} Encoder}
\label{subsec:segenc}
The \texttt{SegFormer} encoder comprises four encoder stages, extracting low to high-level features $\mathbf{f}_1, \mathbf{f}_2, \mathbf{f}_3$, and $\mathbf{f}_4$, as illustrated in \autoref{fig:segenc}. The input image $\VEC{x} \in \mathbb{R}^{H\times W\times 3}$ serves as input to the first stage, where $H$ and $W$ represent the image height and width, respectively. Each stage starts with a two-dimensional convolutional layer, denoted as Conv($k$$\times$$k$, $F_j$)$_{/\rho}$, where $k$$\times$$k$ represents the kernel size, $F_j$ denotes the number of kernels with $j$ indicating the stage index, and $\rho$ representing the stride. This convolutional layer serves as the only dimension-changing layer, thereby functioning as a bottleneck. Subsequently, the convolutional layer is succeeded by layer normalization and a number of $B_j$ transformer blocks. 
The detailed structure of the transformer block is presented in \autoref{fig:transformerblock}. In the first part of the transformer block, an efficient multi-head self-attention (MHSA) operation is applied. To facilitate this operation, the input is flattened in the spatial dimension. Following the attention operation, the result is reshaped back into its original two-dimensional representation. A bypass is employed in parallel to the attention operation. Subsequently, after layer normalization, the Mix-FFN block, 
as introduced by Xie \etal \cite{Xie2021}, is applied, which also has a bypass in parallel to it.
\begin{figure}[t]
\begin{center}
\vspace{3pt}
\resizebox{4.5cm}{!}{
\subimport{./blockdiagrams}{segformer_spatial_reduction.tex}}
\end{center}
   \caption{\textbf{Spatial reduction} $\mathbf{R}(\mathbf{z})$
in the efficient MHSA of \autoref{fig:mhsa}.
   }
\label{fig:spatialreduction}
\end{figure}
The hyperparameters $F_j$ and $B_j$ define the size of the encoder, resulting in the encoder sizes \texttt{MiT-B0}, $\dots$, \texttt{MiT-B5} proposed by Xie \etal \cite{Xie2021}. Note that the \texttt{DynaSegFormer} \cite{Bai} uses the same encoder architecture, but replaces certain matrix multiplications by a specific dynamic gated linear layer (DGL).

\subsection{Efficient Multi-Head Self-Attention (MHSA)}
\label{subsec:mhsa}
The transformer block in the \texttt{SegFormer} encoder (see \autoref{fig:segenc}) employs an efficient scaled dot-product self-attention
\begin{equation}
\label{equ:self-attention-attention-score}
\MAT{H}_i(\VEC{z}) = \textbf{softmax}(\frac{\VEC{z}\VEC{W}_i^\mathrm{(Q)}\cdot(\VEC{R}(\VEC{z})\VEC{W}_i^{\mathrm{(K)}})^{\mathsf{T}}}{\sqrt{F^{\mathrm{out}}}})\cdot\VEC{R}(\VEC{z})\VEC{W}_i^\mathrm{(V)},
\end{equation}
to compute the attention weights $\MAT{H}_i\in\mathbb{R}^{N\times F^\mathrm{out}}$. The learnable fully connected (FC) layer weights are $\VEC{W}_i^\mathrm{(Q)}, \VEC{W}_i^\mathrm{(K)}, \VEC{W}_i^\mathrm{(V)} \in\mathbb{R}^{F^{\mathrm{in}}\times F^{\mathrm{out}}}$, and $()^{\mathsf{T}}$ being the transpose \cite{Dosovitskiy2021}. Here, 
$\VEC{z} = (\VEC{z}_n)^{\mathsf{T}} \in\mathbb{R}^{N\times F^{\mathrm{in}}}$ denotes the input sequence composed of vectors $\VEC{z}_1, \VEC{z}_2, \dots, \VEC{z}_N$ with $\VEC{z}_n\in\mathbb{R}^{F^{\mathrm{in}}}$, where $N = \frac{H}{4\cdot2^{j-1}}\cdot \frac{W}{4\cdot2^{j-1}} \in\mathbb{N}$ represents the number of input vectors, $F^{\mathrm{in}} \in\mathbb{N}$ denotes the length of an input vector, and $F^{\mathrm{out}} \in\mathbb{N}$ represents the length of an output vector. The index $i \in \{1, \dots, I\}$ refers to the head index, with $I \in \mathbb{N}$ being the number of heads. The spatial reduction operation $\mathbf{R}(\VEC{z})$ in (\ref{equ:self-attention-attention-score}), illustrated in \autoref{fig:spatialreduction}, is the only deviation from the standard scaled dot-product self-attention. This operation reduces the dimension of $\VEC{z}$ from $N\times F^{\mathrm{in}}$ to $\frac{N}{(\rho')^2}\times F^{\mathrm{in}}$ using a convolution with kernel size $\rho' \times \rho'$ and stride $\rho'$, where $\rho' \in\mathbb{N}$ is a hyperparameter controlling the reduction rate.

Typically, multiple heads are utilized in parallel, which is then called multi-head self-attention (MHSA) with
\begin{equation*}
\label{equ:msa}
\mathrm{\mathbf{MHSA}}(\VEC{z}) = [
\VEC{H}_1(\VEC{z})\;\;\VEC{H}_2(\VEC{z})\;\;\dots\;\;\VEC{H}_I(\VEC{z})
] \cdot\VEC{W}^\mathrm{MHSA},
\end{equation*}
employing learnable weights $\VEC{W}^\mathrm{MHSA}\in\mathbb{R}^{IF^{\mathrm{out}}\times F^{\mathrm{in}}}$, and $\mathrm{\mathbf{MHSA}}(\VEC{z})\in \mathbb{R}^{N\times F^{\mathrm{in}}}$ \cite{Dosovitskiy2021}.

\hfill
\subsection{\textbf{\texttt{DynaSegFormer}} Specifics and Issues}
\label{subsec:DynaSegFormer}
Bai \etal \cite{Bai} proposed a dynamic structured pruning method, called dynamic gated linear layer (DGL), which can be used to replace expensive matrix multiplications, as, e.g., the ones in an MHSA layer, or in $1\times 1$ convolutions. For the \texttt{DynaSegFormer} \cite{Bai}, the three attention head matrix multiplications $\VEC{z}\VEC{W}_i^\mathrm{(Q)}$, $\VEC{R}(\VEC{z})\VEC{W}_i^\mathrm{(K)}$ and $\VEC{R}(\VEC{z})\VEC{W}_i^\mathrm{(V)}$ in (\ref{equ:self-attention-attention-score}), as well as one of the Mix-FFN $1\times 1$ convolutions, 
introduced by Xie \etal \cite{Xie2021}, all in the transformer block in \autoref{fig:transformerblock}, are replaced by the dynamic gated linear layer (DGL) operation. Within a single multi-head self-attention layer, the DGL operation prunes single dimensions in each attention head, resulting in varying attention head dimensions. This complicates the efficient implementation of the DGL operation, prohibiting it to effectively reduce inference time on a given hardware. Moreover, the dynamic assembly of the learnable fully connected (FC) layer weights $\VEC{W}_i^\mathrm{(Q)}, \VEC{W}_i^\mathrm{(K)}, \VEC{W}_i^\mathrm{(V)}$ during the DGL operation slows down both inference and training, as we observed on \texttt{NVIDIA A100} and \texttt{GTX 1080 Ti} GPUs.

Bai \etal \cite{Bai} further proposed replacing the concatenation operation in the \texttt{SegFormer} decoder with addition. This alteration significantly reduces the computational complexity (FLOPs) of the decoder, while simultaneously enhancing performance (mIoU) on the Cityscapes and ADE20K benchmarks. Additionally, Bai \etal \cite{Bai} utilize feature and logit-based knowledge distillation to further mitigate performance losses in terms of mIoU.
We incorporate the replacement of the costly concatenation by simple addition into our \texttt{StaticSegFormer} decoder, details are shown in \autoref{fig:head_dynasegformer}.

\begin{figure}[t]
\begin{center}
\resizebox{6.cm}{!}{
\subimport{./blockdiagrams}{MHSA_v2.tex}}
\end{center}

   \caption{\textbf{MHSA} with static head pruning in our \textbf{\texttt{StaticSegFormer} encoder}. \textcolor{red}{Red} symbols mark pruning-affected dimensions.}
\label{fig:mhsa}
\end{figure}

\begin{figure}[t!]
\begin{center}
\resizebox{0.75\columnwidth}{!}{
\begin{tikzpicture}[rotate=0,transform shape, line width=1pt,rounded corners=0]

        \node [fill=tu93, shape=rectangle, minimum width=11.1cm, minimum height=7.65cm, align=center] (dec) at (4.2, -2.75) {};
        \node [font=\fontsize{13.0}{13.0}, rotate=0, align=right] (0) at (7.5, -6.0) {\texttt{StaticSegFormer}\\\textbf{Decoder}};
        
         \node[draw, minimum height=1.6em][minimum width=2.5cm, fill=TuGray20] (L1) {Conv(1$\times$1, $F$)};
         \node[draw, minimum height=1.6em][right= 0.1 of L1, minimum width=2.5cm, fill=TuGray20] (L2) {Conv(1$\times$1, $F$)};
         \node[draw, minimum height=1.6em][right= 0.1 of L2, minimum width=2.5cm, fill=TuGray20] (L3) {Conv(1$\times$1, $F$)};
         \node[draw, minimum height=1.6em][right= 0.1of L3, minimum width=2.5cm, fill=TuGray20] (L4) {Conv(1$\times$1, $F$)};

         \node[draw, minimum height=1.6em][below= 0.05 of L2, minimum width=2.5cm, fill=TuGray40] (R2) {Resize};
         \node[draw, minimum height=1.6em][below= 0.05 of L3, minimum width=2.5cm, fill=TuGray40] (R3) {Resize};
         \node[draw, minimum height=1.6em][below= 0.05 of L4, minimum width=2.5cm, fill=TuGray40] (R4) {Resize};
         
         \node[draw, minimum height=1.6em][below right= 1.2cm and -2.538cm of L1, minimum width=10.41cm, fill=tu23] (Concat)      {+};

        \node[draw, minimum height=1.6em][below right= 2.4cm and 1.415cm of L1, minimum width=2.5cm, fill=TuGray20] (L5)      {Conv(1$\times$1, $F$)};
        
         \node[draw, minimum height=1.6em][below right= 3.07cm and 1.415cm of L1, minimum width=2.5cm, fill=TuGray60] (L6)      {Norm};

        \node[draw, minimum height=1.6em][below right= 3.72cm and 1.415cm of L1, minimum width=2.5cm, fill=white] (L7)      {ReLU};

        \node[draw, minimum height=1.6em][below right= 4.92cm and 1.415cm of L1, minimum width=2.5cm, fill=TuGray20] (L8)      {Conv(1$\times$1, $S$)};

        \node[draw, fill=white, minimum height=1.6em][below right= 5.6cm and 1.415cm of L1, minimum width=2.5cm] (L9)      {argmax};
         
      \coordinate (L10) at (3.95, -7.3) {};

	 \coordinate  (inp1) at ($(L1) + (0.0,1.15)$) {};
      \coordinate  (inp2) at ($(L2) + (0.0,1.15)$) {};
      \coordinate  (inp3) at ($(L3) + (0.0,1.15)$) {};
      \coordinate  (inp4) at ($(L4) + (0.0,1.15)$) {};
	 \node[label, rotate=0, font=\fontsize{13.4}{13.4}] (inp1text) [above left= 0.0 and -1.3 of L1] {$\mathbf{f}_1$};
	 \node[label, rotate=0, font=\fontsize{13.4}{13.4}] (inp2text) [above left= 0.0 and -1.3 of L2] {$\mathbf{f}_2$};
	 \node[label, rotate=0, font=\fontsize{13.4}{13.4}] (inp3text) [above left= 0.0 and -1.3 of L3] {$\mathbf{f}_3$};
	 \node[label, rotate=0, font=\fontsize{13.4}{13.4}] (inp4text) [above left= 0.0 and -1.3 of L4] {$\mathbf{f}_4$};

      \draw [-{Triangle[length=1.75mm,width=1.75mm]}] (inp1) -- (L1);
      \draw [-{Triangle[length=1.75mm,width=1.75mm]}] (inp2) -- (L2);
      \draw [-{Triangle[length=1.75mm,width=1.75mm]}] (inp3) -- (L3);
      \draw [-{Triangle[length=1.75mm,width=1.75mm]}] (inp4) -- (L4);


      \draw [-{Triangle[length=1.75mm,width=1.75mm]}] (L1) -- (0, -1.53);
      \draw [-{Triangle[length=1.75mm,width=1.75mm]}] (R2) -- (2.63, -1.53);
      \draw [-{Triangle[length=1.75mm,width=1.75mm]}] (R3) -- (5.265, -1.53);
      \draw [-{Triangle[length=1.75mm,width=1.75mm]}] (R4) -- (7.905, -1.53);

      \draw [-{Triangle[length=1.75mm,width=1.75mm]}] (Concat) -- (L5);
      \draw [-{Triangle[length=1.75mm,width=1.75mm]}] (L7) -- (L8);
      \draw [-{Triangle[length=1.75mm,width=1.75mm]}] (L9) -- (L10);


      \node [rotate=0] (f1dim) at (0.89, 0.75) {$\frac{H}{4}$$\times$$\frac{W}{4}$$\times$$F_1$};

      \node [rotate=0] (f1dim) at (3.52, 0.75) {$\frac{H}{8}$$\times$$\frac{W}{8}$$\times$$F_2$};

      \node [rotate=0] (f1dim) at (6.17, 0.75) {$\frac{H}{16}$$\times$$\frac{W}{16}$$\times$$F_3$};

      \node [rotate=0] (f1dim) at (8.805, 0.75) {$\frac{H}{32}$$\times$$\frac{W}{32}$$\times$$F_4$};

      \node [rotate=0] (f1dim) at (0.83, -1.21) {$\frac{H}{4}$$\times$$\frac{W}{4}$$\times$$F$};
      \node [rotate=0] (f1dim) at (3.45, -1.21) {$\frac{H}{4}$$\times$$\frac{W}{4}$$\times$$F$};
      \node [rotate=0] (f1dim) at (6.10, -1.21) {$\frac{H}{4}$$\times$$\frac{W}{4}$$\times$$F$};
      \node [rotate=0] (f1dim) at (8.72, -1.21) {$\frac{H}{4}$$\times$$\frac{W}{4}$$\times$$F$};

      \node [rotate=0] (f1dim) at (4.77, -2.4) {$\frac{H}{4}$$\times$$\frac{W}{4}$$\times$$F$};
      \node [rotate=0] (f1dim) at (4.77, -4.9) {$\frac{H}{4}$$\times$$\frac{W}{4}$$\times$$F$};

      \node [rotate=0] (f1dim) at (4.77, -6.85) {$\frac{H}{4}$$\times$$\frac{W}{4}$$\times$$S$};

      \node [rotate=0, font=\fontsize{13.4}{13.4}, outer sep= 0em, inner sep= 0.25mm] (f1dim) at (3.7, -7.2) {$\mathbf{y}$};

\end{tikzpicture}}
\end{center}

   \caption{\textbf{Our \texttt{StaticSegFormer} decoder},
based on \texttt{DynaSegFormer}~\cite{Bai}.}
   
\label{fig:head_dynasegformer}
\end{figure}

\subsection{Our Novel Encoder Head Pruning}
\label{subsec:head_prunig}
Due to the limitations associated with the dynamic pruning method employed in the \texttt{DynaSegFormer} network, we propose to prune the number $I$ of heads prior to fine-tuning{---}\textit{in a static fashion}. Our modified multi-head self-attention block is shown in \autoref{fig:mhsa}. We introduce the global pruning ratio $r \in \I$, with $\I = [0,1]$. The final number of heads is then obtained by

\begin{equation}
    I^\mathrm{prune} = \lfloor r\cdot I\rfloor \leq I,
\end{equation}
with $\lfloor \cdot \rfloor$ being the flooring operator. Additionally, to remove one head, the corresponding weights in $\VEC{W}^\mathrm{MHSA}$ must also be pruned. Consequently, we reduce the dimension of $\VEC{W}^\mathrm{MHSA}$ from $IF^{\mathrm{out}}\times F^{\mathrm{in}}$ to $I^\mathrm{prune}F^{\mathrm{out}}\times F^{\mathrm{in}}$, resulting in the weight matrix $\VEC{W}^\mathrm{MHSA, prune} \in\mathbb{R}^{I^\mathrm{prune}F^{\mathrm{out}}\times F^{\mathrm{in}}}$. Therefore, the only entities that change are the number of heads $I^\mathrm{prune}$ and $\VEC{W}^\mathrm{MHSA, prune}$, while all dimensions and weights inside each head remain unchanged. \textit{Although this approach at first glance looks like a minor change in the pruning strategy, we will show its significant effect on achieving a computationally efficient semantic segmentation of still high performance.}

For the unpruned multi-head self-attention of our baseline model, the output dimensions $F^{\mathrm{out}}$ of the $I$ attention heads sum up to $F^{\mathrm{in}}$, as it holds $F^{\mathrm{out}} \cdot I = F^{\mathrm{in}}$. Our proposed pruning method removes complete attention heads, including their corresponding learnable FC layer weights $\VEC{W}_i^\mathrm{(Q)}, \VEC{W}_i^\mathrm{(K)}, \VEC{W}_i^\mathrm{(V)}$. Nevertheless, the inner dimensions of the remaining attention heads remain unchanged, as shown in \autoref{fig:mhsa}. Therefore, $F^{\mathrm{in}}$ and $F^{\mathrm{out}}$ remain unchanged, and the (after pruning) remaining learnable FC layer weights $\VEC{W}_i^\mathrm{(Q)}, \VEC{W}_i^\mathrm{(K)}$ and $\VEC{W}_i^\mathrm{(V)}$ maintain the same dimensions $F^{\mathrm{in}}\times F^{\mathrm{out}}$. However, $I$ is reduced to $I^\mathrm{prune}$, so the output dimensions $F^{\mathrm{out}}$ of the remaining $I^\mathrm{prune}$ attention heads no longer sum up to $F^{\mathrm{in}}$, as $F^{\mathrm{out}}\cdot I^\mathrm{prune} < F^{\mathrm{in}}$. Consequently, by pruning entire attention heads while keeping $F^{\mathrm{out}}$ and $F^{\mathrm{in}}$ unchanged, we introduce a bottleneck into the MHSA layer. Therefore, the expensive matrix multiplications of key, query, and value inside the pruned multi-head self-attention are performed on a compressed input representation, \textit{which proves to be advantageous as it reduces the computational complexity (FLOPs) immensely, while maintaining high mIoU performance.}

Unlike the dynamic structured pruning of \texttt{DynaSeg-} \texttt{Former},
our static encoder pruning reduces training time, enabling
extensive ablations of pruning ratio $r$ across model sizes.
For the decoder, we adopt the modifications of Bai \etal \cite{Bai},
which reduce FLOPs without sacrificing mIoU.

\begin{table}[t]
\centering
  \caption{\textbf{Datasets and splits used in our experiments}. Training and validation refer to the official splits proposed from the respective paper, while train* and val* refer to custom splits employed in this work for our ablation study.}
  
    \setlength{\tabcolsep}{.25em}
    \begingroup
    \renewcommand{\arraystretch}{0.35}
    \begin{tabular}{@{}llrl@{}}\toprule
       Dataset & Splits & \# Images & Symbol\\
        \midrule
       
        \multirow{5}{*}{Cityscapes~\cite{Cordts2016}}& train &  2,975 & $\cstrain$ \\[3pt]
        
        & $\hookrightarrow$ train* &  2,475 & $\cstrainstar$ \\[3pt]
        & $\hookrightarrow$ val* &  500 & $\csvalstar$\\[3pt]
        & val &  500 & $\csval$ \\[3pt]
        \midrule
        \multirow{2}{*}{ADE20K~\cite{Zhou2017}}& train & 25,574 & $\adetrain$ \\[3pt]
        & val & 2,000 & $\adeval$ \\[3pt]
        \bottomrule     
    \end{tabular}
    \endgroup

  \label{tab:datasets}
\end{table}
\section{Evaluation Setup}
\subsection{Dataset}
Following the approach of Bai \etal \cite{Bai}, we evaluate our pruning method on the Cityscapes~\cite{Cordts2016} and ADE20K~\cite{Zhou2017} datasets. Details regarding the employed splits and hyperparameters are presented in \autoref{tab:datasets} and \autoref{table:hyperparameters}, respectively. Additionally, for all experiments, we utilize ImageNet~\cite{Deng2009} pretrained weights for weight initialization from the \texttt{MMSegmentation} toolbox~\cite{mmseg2020}.

\textbf{Cityscapes}: We utilize the Cityscapes benchmark~\cite{Cordts2016} as the primary dataset for our ablation study. Following common practice, also employed by Bai \etal \cite{Bai}, we present our final results on the Cityscapes validation split, denoted as $\csval$. To avoid hyperparameter tuning on the $\csval$ split, which we reserve for our final comparison, we further split the training data $\cstrain$ into $\cstrainstar$ and $\csvalstar$ for our ablation study. All decision-making was conducted solely based on the $\csvalstar$ split. We retrained the best models on the full $\cstrain$ split and report results for the $\csval$ in \autoref{fig:results_cs}, \autoref{fig:miou_vs_fps_cs}, and in \autoref{tab:data_augmentations}.

\textbf{ADE20K}: Since the ADE20K \cite{Zhou2017} semantic segmentation benchmark has been utilized in many previous works and presents a greater challenge for structured pruning due to the higher number of classes, we also present results for it in \autoref{tab:data_augmentations}. The best performing models, based on the results for the $\csvalstar$ split, are retrained on the original training split $\adetrain$ and evaluated on the original validation split $\adeval$.
\subsection{Implementation Details}
For all our experiments, we utilized the \texttt{MMSegmentation} toolbox~\cite{mmseg2020}
and for the baseline experiments its included implementation for the \texttt{SegFormer} \cite{Xie2021} with standard hyperparameter settings. Additionally, for all \texttt{StaticSegFormer} experiments, we extended it by our employed decoder
and the capability to prune the MHSA block in the encoder.
We employed the AdamW \cite{Loshchilov2019} optimizer and utilized polynomial learning rate scheduling. Unless stated otherwise, experiments were conducted on an \texttt{NVIDIA A100} GPU. During training, we used random cropping with an input resolution of $1024 \times 1024$ for Cityscapes and $512 \times 512$ for ADE20K. More details on the hyperparameters can be found in \autoref{table:hyperparameters}. We maintained consistent settings across all model types, regardless of the network architecture. For ADE20K we employed a batch size of 16, whereas for Cityscapes, we used 8. Furthermore, the training resolution used during cropping differs and is $1024\times 1024$ for Cityscapes and $512\times 512$ for ADE20K. Additionally, the number of iterations is 80,000 for Cityscapes, while for ADE20K, we used 160,000. All other hyperparameters remained identical for the datasets utilized.

For evaluation, we used the mIoU metric, commonly utilized in semantic segmentation tasks. Additionally, we measured efficiency using a resolution of $2048 \times 1024$ for Cityscapes and report the metrics floating-point operations (FLOPs) and frames per second [fps]. The FLOPs were measured using random number tensors with sizes $1024\times 2048 \times 3$. The frame rate was measured on an \texttt{NVIDIA A100} and a \texttt{GTX 1080 Ti} GPU by averaging the inference time over 200 samples after 200 warmup iterations.
\begin{table}[t]
  \caption{\textbf{Hyperparameters used for training} the \texttt{SegFormer} \texttt{DynaSegFormer} and our proposed \texttt{StaticSegFormer} on the Cityscapes and ADE20K datasets.}
  \label{tab:headings}
  \centering
  \begingroup
  \renewcommand{\arraystretch}{0.65}
  \begin{tabular}{@{}lcc@{}}
    \toprule
     \centering \textbf{Hyperparameters} &  Cityscapes & ADE20K  \\
    \midrule
     Batch size   & 8 & 16  \\ 

     Random Crop & $1024\times 1024$ & $512\times 512$ \\
     
     $\#$ of training iterations  & 80,000  & 160,000  \\ 

     Initial learning rate encoder & $6\cdot 10^{-5}$ & $6\cdot 10^{-5}$  \\
     Initial learning rate decoder & $6\cdot 10^{-4}$ & $6\cdot 10^{-4}$  \\

     Learning rate schedule   & polynomial  & polynomial   \\ 

     Optimizer & AdamW & AdamW  \\ 
     Optimizer parameters $\beta_{1}$, $\beta_{2}$   & 0.9, 0.999 & 0.9, 0.999  \\ 

     Weight decay   & 0.01 & 0.01  \\ 
  \bottomrule
  \end{tabular}\endgroup
  
\label{table:hyperparameters}
\end{table}
\section{Results and Discussion}

\subsection{Comparison with Baseline Methods}
We present results regarding mIoU and FLOPs for Cityscapes in \autoref{fig:results_cs}. We plot all models lying on the Pareto frontier of the ablation study, meaning there is no other model with higher mIoU and lower FLOPs. Additionally, we compare the frame rate of our \texttt{StaticSegFormer} with the \texttt{DynaSegFormer} and \texttt{SegFormer} for Cityscapes in \autoref{fig:miou_vs_fps_cs}. Detailed numbers of the models are reported in \autoref{tab:data_augmentations}. 
\begin{figure}[t]
\begin{center}
\subimport{./figures}{flops_miou_segformer_best_v2.tex}
\end{center}

   \caption{\textbf{mIoU vs.\ FLOPs} of our \texttt{StaticSegFormer},
\texttt{DynaSegFormer} \cite{Bai}, and baseline
\texttt{SegFormer} \cite{Xie2021} on
$\mathcal{D}^\mathrm{CS}_\mathrm{val}$.
All models are trained on $\mathcal{D}^\mathrm{CS}_\mathrm{train}$.
* denotes results from the respective papers.
FLOPs for $2048 \times 1024$.}
\label{fig:results_cs}
\end{figure}
\begin{figure}[t]
\begin{center}
\subimport{./figures}{fps_miou_segformer_best_cs.tex}
\end{center}

   \caption{\textbf{mIoU vs.\ frame rate [fps]} of our
\texttt{StaticSegFormer}, \texttt{DynaSegFormer} \cite{Bai},
and baseline \texttt{SegFormer} \cite{Xie2021} on
$\mathcal{D}^\mathrm{CS}_\mathrm{val}$.
All models are trained on $\mathcal{D}^\mathrm{CS}_\mathrm{train}$.
* denotes mIoU values from the respective papers.
Frame rates for $2048 \times 1024$ on an \texttt{NVIDIA A100} GPU.}
   
\label{fig:miou_vs_fps_cs}
\end{figure}

\begin{table*}[t]
  \caption{\textbf{Performance of our \texttt{StaticSegFormer}}
vs.\ \texttt{SegFormer} \cite{Xie2021} and
\texttt{DynaSegFormer} \cite{Bai} on
$\mathcal{D}^\mathrm{CS}_\mathrm{val}$ and
$\mathcal{D}^\mathrm{ADE}_\mathrm{val}$.
Models are trained on the respective training splits.
Results are grouped by FLOPs, measured at $2048\times1024$. * denotes values from the respective papers.
Frame rates are measured on \texttt{NVIDIA A100} and
\texttt{GTX 1080 Ti} GPUs.}

\extrarowheight=\aboverulesep
    \addtolength{\extrarowheight}{\belowrulesep}
    \aboverulesep=0pt
    \belowrulesep=0pt
    \resizebox{\textwidth}{!}{\begingroup
\renewcommand{\arraystretch}{0.65}\begin{tabular}{@{}llccrcccc@{}}
    \toprule[.9pt]
    \rule{0mm}{3.5ex}\textbf{Pruning Method} & \textbf{Network}&\textbf{\makecell{ Pruning \\[-2pt] Ratio $r$}} & \makecell{\textbf{\# Params}\\\textbf{(M)}} & \textbf{\makecell{FLOPs \\(G)}} & \textbf{\makecell{Frame Rate\\[-2pt] [fps] A100}} & \textbf{\makecell{Frame Rate\\[-2pt] [fps] 1080 Ti}} & \textbf{\makecell{mIoU $\csval$ \\(\%)}} & \textbf{\makecell{mIoU $\mathcal{D}^\mathrm{ADE}_\mathrm{val}$ \\(\%)}}\\
    \midrule
    \texttt{DynaSegFormer}* \cite{Bai} & \texttt{SegFormer MiT-B0} & 0.5 & 4.2 & \underline{47} & 1.4 & 1.4 & 74.3 & 35.0 \\
    \texttt{DynaSegFormer}* \cite{Bai} & \texttt{SegFormer MiT-B0} & 0.7 & 4.2 & 63 & 2.5 & 0.9 & \underline{75.1} & \textbf{36.9}\\
    Ours: \texttt{StaticSegFormer} & \texttt{SegFormer MiT-B0} & 0.1 & \textbf{2.8} & \textbf{33} & \textbf{35.7} & \textbf{7.5} & 73.8 & 33.1 \\
    Ours: \texttt{StaticSegFormer} & \texttt{SegFormer MiT-B0} & 0.5 & \underline{3.1} & 60 & \underline{31.7} & \textbf{7.5} & \textbf{76.2} & \underline{35.3} \\
    \midrule
    No Pruning (baseline) & \texttt{SegFormer MiT-B0} & - & \textbf{3.8} & \underline{126} & \underline{23.6}& \textbf{5.7} & \underline{76.2} & \underline{37.4} \\
    Ours: \texttt{StaticSegFormer} & \texttt{SegFormer MiT-B1} & 0.3 & \underline{11.1} & \textbf{109} & \textbf{25.1} & \underline{5.1} & \textbf{76.7} & \textbf{38.6} \\
    \midrule[.9pt]
    No Pruning (baseline) & \texttt{SegFormer MiT-B1} & - & \textbf{13.7} & 244 & \textbf{17.4} & \textbf{3.8} & 78.5 & 42.2 \\
    \texttt{DynaSegFormer}* \cite{Bai} & \texttt{SegFormer MiT-B2} & 0.7 & 30.9 & 287 & 0.2 & 0.2 & \underline{80.3} & \underline{44.6} \\
    \texttt{DynaSegFormer}* \cite{Bai} & \texttt{SegFormer MiT-B2} & 0.5 & 30.9 & \underline{214} & 0.3 & 0.2 & 79.8 & \textbf{45.4} \\
    Ours: \texttt{StaticSegFormer} & \texttt{SegFormer MiT-B2} & 0.3 & \underline{20.0} & \textbf{192} & \underline{15.9} & \underline{3.2} & \textbf{80.5} & 43.2 \\
    \midrule[.9pt]
    No Pruning (baseline) & \texttt{SegFormer MiT-B2} & - & \textbf{27.5} & \underline{717} & \textbf{10.1} & \textbf{2.5} & \underline{81.0} & \underline{46.5} \\
    Ours: \texttt{StaticSegFormer} & \texttt{SegFormer MiT-B3} & 0.7 & \underline{40.0} & \textbf{471} & \underline{9.2} & \underline{2.2} & \textbf{81.7} & \textbf{46.9} \\
    \midrule[.9pt]
    No Pruning (baseline) & \texttt{SegFormer MiT-B3} & - & \textbf{47.3} & \underline{963} & \textbf{6.8} & \textbf{1.7} & \underline{81.7} & \textbf{49.5} \\
    Ours: \texttt{StaticSegFormer} & \texttt{SegFormer MiT-B4} & 0.9 & \underline{50.0} & \textbf{762} & \underline{5.8} & \underline{1.5} & \textbf{82.2} & \underline{47.9} \\
    \bottomrule[.9pt]
\end{tabular}\endgroup}

  \label{tab:data_augmentations}
  
\end{table*}

In \autoref{fig:results_cs} we show the mIoU performance and FLOPs of our proposed \texttt{StaticSegFormer} against the baseline \texttt{SegFormer} \cite{Xie2021} and the \texttt{DynaSegFormer} \cite{Bai} on $\mathcal{D}^\mathrm{CS}_\mathrm{val}$ for the various encoder sizes \texttt{MiT-B}$n, n\in\{0,1,\dots, 5\}$, represented by different marker colors. For the \texttt{DynaSegFormer} \cite{Bai}, the values are taken from the respective paper and use a pruning ratio of $r=0.7$ and $r=0.5$ for both \texttt{MiT-B0} and \texttt{B2}. Based on the results of our ablation study, we selected $r \in \{0.1, 0.5, 0.7\}$ for \texttt{MiT-B0}, $r \in \{0.1, 0.3\}$ for \texttt{MiT-B1}, $r \in \{0.1, 0.3, 0.5\}$ for \texttt{MiT-B2}, $r \in \{0.1, 0.3, 0.5, 0.7\}$ for \texttt{MiT-B3}, $r \in \{0.9\}$ for \texttt{MiT-B4}, and $r \in \{0.1\}$ for \texttt{MiT-B5}.

For very low FLOPs, our simpler \texttt{StaticSegFormer} with the \texttt{MiT-B0} encoder performs as well as the \texttt{DynaSegFormer}. For all other model sizes ($n \in {1,2,3,4,5}$) and pruning ratios, our \texttt{StaticSegFormer} outperforms the \texttt{DynaSegFormer}. Compared to the baseline, we achieve significant improvements for all encoder sizes. As an example, the \texttt{StaticSegFormer MiT-B3} with $r=0.1$ outperforms the baseline \texttt{SegFormer MiT-B1} by a $2.75\%$ absolute higher mIoU performance at a similar computational complexity. In general, we observe that \textit{using a highly pruned large encoder is advantageous compared to a smaller unpruned one}. Also computational complexity savings can be obtained by the \texttt{StaticSegFormer} vs.\ \texttt{SegFormer} while preserving mIoU: Our pruned \texttt{MiT-B3} ($r=0.7$) requires only about 50\% of the computations of the baseline \texttt{MiT-B3}, whereas $1/3$ of complexity savings are possible with \texttt{MiT-B3} ($r=0.9$) against the larger \texttt{MiT-B4} baseline.

In \autoref{fig:miou_vs_fps_cs}, we report the mIoU vs.\ the frame rate on an \texttt{NVIDIA A100} GPU on $\mathcal{D}^\mathrm{CS}_\mathrm{val}$ for the same models as in \autoref{fig:results_cs}. We again report results for the baseline \texttt{SegFormer} \cite{Xie2021} and for the two pruning methods \texttt{DynaSegFormer} \cite{Bai} and \texttt{StaticSegFormer}.

We observe that the \texttt{DynaSegFormer} \textit{drastically reduces the frame rate} compared to the baseline \texttt{SegFormer}, regardless of the encoder size or pruning ratio, making it highly impractical for deployment on this GPU platform. The \texttt{StaticSegFormer MiT-B0} ($r=0.7$), on the other hand, has better mIoU performance than the \texttt{SegFormer MiT-B0}, but achieves a 7 fps higher frame rate. The \texttt{StaticSegFormer MiT-B3} ($r=0.7$) outperforms the \texttt{SegFormer MiT-B3} by a higher frame rate (9 fps instead of 7 fps), while achieving similar mIoU performance. To conclude, the \texttt{StaticSegFormer} \textit{allows a higher frame rate for all encoder sizes, exceeding the baseline in most cases and the reference method} \texttt{DynaSegFormer} \textit{by far}.

In \autoref{tab:data_augmentations}, we show further details. Based on FLOPs, we assign the methods to five groups and report the mIoU performance for $\mathcal{D}^\mathrm{CS}_\mathrm{val}$ and $\mathcal{D}^\mathrm{ADE}_\mathrm{val}$, the number of parameters, global pruning ratios, and frame rates on an \texttt{NVIDIA A100} and on a \texttt{GTX 1080 Ti} GPU. The values for \texttt{DynaSegFormer} are taken from Bai \etal \cite{Bai}.

Concerning mIoU on $\mathcal{D}^\mathrm{CS}_\mathrm{val}$, our \texttt{StaticSegFormer MiT-B0} clearly outperforms both the \texttt{DynaSegFormer} and the baseline \texttt{SegFormer}. Specifically, in the lowest complexity group (upper table segment), the \texttt{StaticSegFormer MiT-B0} with $r=0.5$ requires fewer FLOPs (60 GFLOPs) compared to \texttt{DynaSegFormer MiT-B0} with $r=0.7$ (63 GFLOPs), achieving a 1.1\% absolute higher mIoU (76.2\% vs.\ 75.1\%) \textit{while operating at frame rates 12.7 and 8.3 times higher on an \texttt{NVIDIA A100} and a \texttt{GTX 1080 Ti} GPU, respectively}. The \texttt{DynaSegFormer} uses dynamic structured pruning and therefore increases the number of parameters compared to the baseline \texttt{SegFormer}, while the \texttt{StaticSegFormer} achieves a reduction in parameters of 18.4\% relative to the baseline \texttt{MiT-B0}. In all four reported cases (\texttt{MiT-B0} and \texttt{MiT-B2}, $r=0.5$ and $0.7$), \texttt{DynaSegFormer} shows inacceptably low frame rates both on the \texttt{NVIDIA A100} and \texttt{GTX 1080 Ti} GPU, making it impractical for usage on a GPU platform. In contrast, the \texttt{StaticSegFormer MiT-B0} ($r=0.5$) \textit{increases frame rates vs.\ the baseline }\texttt{MiT-B0}\textit{ by 34.3\% relative} (\texttt{NVIDIA A100}) \textit{and 31.5\% relative} (\texttt{GTX 1080 Ti}) and \textit{reduces FLOPs to less than 50\%}, \textit{while having no mIoU performance drop on $\mathcal{D}^\mathrm{CS}_\mathrm{val}$ at all}. In the middle complexity group, the \texttt{StaticSegFormer MiT-B2} with $r=0.3$ surpasses both \texttt{DynaSegFormer MiT-B2} models in all reported metrics for the $\mathcal{D}^\mathrm{CS}_\mathrm{val}$ on both GPU platforms.
Regarding mIoU on $\mathcal{D}^\mathrm{ADE}_\mathrm{val}$, we observe that while our \texttt{StaticSegFormer} outperforms the baseline \texttt{SegFormer} in most cases, it falls short compared to \texttt{DynaSegFormer}. Specifically, the \texttt{DynaSegFormer} achieves higher mIoU values, likely benefiting from its dynamic structured pruning, which allows for better adaptation to the increased complexity of the ADE20k dataset with its larger number of classes. However, the frame rates of \texttt{DynaSegFormer} remain impractically low on both the \texttt{NVIDIA A100} and \texttt{GTX 1080 Ti}, further emphasizing its limited applicability in real-time scenarios.

Across all complexity groups, \texttt{StaticSegFormer}
achieves the highest mIoU on $\mathcal{D}^\mathrm{CS}_\mathrm{val}$
and the lowest FLOPs, with frame rates close to the baseline
\texttt{SegFormer} on both GPUs. For example, the \texttt{StaticSegFormer MiT-B3} with $r=0.7$ surpasses the baseline \texttt{SegFormer MiT-B2} with 34.3\% relative fewer FLOPs and a 0.7\% absolute higher mIoU on $\mathcal{D}^\mathrm{CS}_\mathrm{val}$. We note that often the baseline method turns out to be a bit better than our approach in frame rate, however, at costs both in mIoU and FLOPs.

Overall, across all encoder sizes, our \texttt{StaticSegFormer}
achieves a better FLOPs–frame rate trade-off and higher mIoU
than \texttt{SegFormer}, while offering attractive frame rates
compared to \texttt{DynaSegFormer}.

\subsection{Ablation Study}
We conducted an ablation study on the pruning ratio $r$ over five values, $r \in \{0.1, 0.3, 0.5, 0.7, 0.9\}$, for each of the six encoder sizes \texttt{MiT-B0}, $\dots$, \texttt{MiT-B5} on the Cityscapes custom validation split $\csvalstar$, see \autoref{tab:datasets}, as shown in \autoref{fig:ablationstudy_csdev}. We report computational complexity, measured in FLOPs for a resolution of $2048\times1024$, and model performance (mIoU). The different encoder sizes are represented as different colors. The triangular markers show the results for our \texttt{StaticSegFormer} for different global pruning ratios, where the leftmost point for each encoder size always refers to $r=0.1$ and the rightmost to $r=0.9$.

The study shows consistent behavior across all model sizes and pruning ratios, with minor statistical fluctuations. Most data points lie above the baseline \texttt{SegFormer} curve, demonstrating the superiority of our approach in reducing FLOPs. For example, the \texttt{StaticSegFormer MiT-B2} ($r=0.3$) requires fewer FLOPs than the baseline \texttt{SegFormer MiT-B1}, yet achieves much higher mIoU performance. Furthermore, this \texttt{StaticSegFormer} model achieves mIoU performance similar to the unpruned \texttt{SegFormer MiT-B2} baseline while operating at less than one-third of its computational complexity.  Concluding, \textit{we observe that it is advantageous to use a larger, heavily pruned encoder rather than a smaller, unpruned one}.

\begin{figure}[t]
\begin{center}
\vspace{3pt}
\small
\begin{tikzpicture}
\pgfplotstableread[col sep=comma]{figures/csv_data/flops_miou_segformer_cs_dev_real.csv}\datatable

\begin{axis}[
name=plot3,
at={(0cm,0cm)},
width=86mm,
height=60mm,
legend cell align={left},
axis line style={tu7},
grid style={tu73},
legend style={fill opacity=1.0, draw opacity=1, text opacity=1, at={(1.03,0.97)}, anchor=south east, draw=tu73, fill=white, line width=0.85,}, 
tick align=outside,
tick pos=left,
xmin=0, xmax=1510,
xlabel={FLOPs (G)},
xmajorgrids,
ytick={72,74,76,78,80,82},
ylabel={mIoU on $\mathcal{D}^\mathrm{CS}_\mathrm{val*}$ (\%)},
ylabel shift = 0mm,
ymajorgrids,
ymin=72, ymax=84
]
\addplot[
      scatter,
      scatter src=explicit symbolic,
      densely dashed,
      mark options={solid},
      mark size=2.25pt,
      line width=1.0pt,
      point meta=explicit symbolic,
      scatter/classes={0={mark=o,tu1},
                   1={mark=o,tu2},
                   2={mark=o,tu3},
                   3={mark=o,tu4},
                   4={mark=o,tu72},
                   5={mark=o,tu8}
      },
      x filter/.expression={\thisrow{method_num} == 0 ? x : NaN},
      visualization depends on={value \thisrow{meta_class} \as \pointclass},
                  ] table [x=Flops, y=mIoU, meta=meta_class, col sep=comma] {\datatable};
\label{baseline6}



\addplot[
      scatter,
      scatter src=explicit symbolic,
      point meta=explicit symbolic,
      mark size=2.25pt,
      line width=1.0pt,
      scatter/classes={20={mark=triangle,tu1},
                   21={mark=triangle,tu2},
                   22={mark=triangle,tu3},
                   23={mark=triangle,tu4},
                   24={mark=triangle,tu72},
                   25={mark=triangle,tu8}
      },
      x filter/.expression={\thisrow{method_num} == 2 ? x : NaN},
      visualization depends on={value \thisrow{meta_class} \as \pointclass},
                  ] table [only marks, x=Flops, y=mIoU, meta=meta_class, col sep=comma] {\datatable};
\label{staticsegformerA6}


\addlegendimage{mark=o, densely dashed, mark options={solid}, mark size=2.25, black,line width=1pt}
\label{baseline_dev}
\addlegendimage{mark=triangle, mark size=2.25, white, mark options={black},line width=1.0pt}
\label{staticsegformer_dev}

\addlegendimage{no marks, tu1,line width=1.0pt, shorten <=5pt, shorten >=5pt}
\label{MiT-B0-A6}
\addlegendimage{no marks, tu2,line width=1.0pt, shorten <=5pt, shorten >=5pt}
\label{MiT-B1-A6}
\addlegendimage{no marks, tu3,line width=1.0pt, shorten <=5pt, shorten >=5pt}
\label{MiT-B2-A6}
\addlegendimage{no marks, tu4,line width=1.0pt, shorten <=5pt, shorten >=5pt}
\label{MiT-B3-A6}
\addlegendimage{no marks, tu72,line width=1.0pt, shorten <=5pt, shorten >=5pt}
\label{MiT-B4-A6}
\addlegendimage{no marks, tu8,line width=1.0pt, shorten <=5pt, shorten >=5pt}
\label{MiT-B5-A6}

\node[right, tu3] at (axis cs:00,82.3) {r=0.3};
\draw[] (axis cs:170, 81.3) -- (axis cs:120,81.9);



\node [draw=tu71,fill=white,fill opacity=01,inner xsep=2,minimum height = 7mm] at (rel axis cs: 0.38,0.142) {\shortstack[l]{
\hspace*{15mm}\textbf{Method}\\
\remHyp{\ref{staticsegformer_dev}} Ours: \texttt{StaticSegFormer}\\[-3pt]
\remHyp{\ref{baseline_dev}} Baseline: \texttt{SegFormer} \cite{Xie2021}\\[-3pt]}};

\node [draw=tu71,fill=white,fill opacity=01,inner xsep=-2,minimum height = 7mm] at (rel axis cs: 0.85,0.285) {\shortstack[l]{
\hspace*{4.5mm}\textbf{Encoder}\hspace*{4.5mm}\\
\remHyp{\ref{MiT-B5-A6}} \texttt{MiT-B5}\\[-3pt]\\
\remHyp{\ref{MiT-B4-A6}} \texttt{MiT-B4}\\[-3pt]\\
\remHyp{\ref{MiT-B3-A6}} \texttt{MiT-B3}\\[-3pt]\\
\remHyp{\ref{MiT-B2-A6}} \texttt{MiT-B2}\\[-3pt]\\
\remHyp{\ref{MiT-B1-A6}} \texttt{MiT-B1}\\[-3pt]\\
\remHyp{\ref{MiT-B0-A6}} \texttt{MiT-B0}\\[-3pt]}};
 
\end{axis}
\end{tikzpicture}
\end{center}

   \caption{\textbf{Ablation of MHSA pruning ratio} $r$
\textbf{across \texttt{SegFormer} encoders} on
$\mathcal{D}^\mathrm{CS}_\mathrm{val*}$.
All models are trained on $\mathcal{D}^\mathrm{CS}_\mathrm{train*}$.
Global ratios $r \in \{0.1,0.3,0.5,0.7,0.9\}$ range
from the lowest/leftmost ($r=0.1$) to the highest/rightmost
($r=0.9$) point for each encoder.
FLOPs for $2048 \times 1024$.
Baseline results are reproduced.}
\label{fig:ablationstudy_csdev}

\end{figure}

\section{Conclusions}
In this work, we demonstrate that the efficiency of semantic segmentation can be significantly improved through our proposed static structured pruning method for attention heads. While structured pruning methods primarily reduce computational complexity (FLOPs), we observed that  dynamic structured pruning can cause very low frame rates on typical GPU target hardware. Our proposed \texttt{StaticSegFormer} successfully decreases FLOPs by up to 50\% and increases the frame rate by up to 34\%, without any loss in mIoU performance on the Cityscapes dataset. Moreover, we observe that employing a larger, strongly pruned model often outperforms a smaller, unpruned one.

{\small
\bibliographystyle{ieee_fullname}
\bibliography{egbib}
}

\end{document}